\documentclass[11pt]{article}

\usepackage[final]{acl}

\usepackage{times}
\usepackage{latexsym}

\usepackage[T1]{fontenc}

\usepackage[utf8]{inputenc}

\usepackage{microtype}

\usepackage{inconsolata}

\usepackage{graphicx}

\usepackage{booktabs}

\usepackage{listings}
\title{Temporal Graph Network: Hallucination Detection in Multi-Turn Conversation}

\author{
  Vidhi Rathore  \quad
  Sambu Aneesh  \quad
  Himanshu Singh  \\
  IIIT Hyderabad \\
  \texttt{\{vidhi.rathore, sambu.aneesh, himanshu.s\}@research.iiit.ac.in}
}

\begin{document}
\maketitle
\begin{abstract}
Hallucinations can be produced by conversational AI systems, particularly in multi-turn conversations where context changes and contradictions may eventually surface.  By representing the entire conversation as a temporal graph, we present a novel graph-based method for detecting dialogue-level hallucinations. Our framework models each dialogue as a node, encoding it using a sentence transformer. We explore two different ways of connectivity: i) shared-entity edges, which connect turns that refer to the same entities; ii) temporal edges, which connect contiguous turns in the conversation. Message-passing is used to update the node embeddings, allowing flow of information between related nodes. The context-aware node embeddings are then combined using attention pooling into a single vector, which is then passed on to a classifier to determine the presence and type of hallucinations.  We demonstrate that our method offers slightly improved performance over existing methods. Further, we show the attention mechanism can be used to justify the decision making process. The code and model weights are made available at: \url{https://github.com/sambuaneesh/anlp-project}.
\end{abstract}

\section{Introduction}

Human-machine interaction has been transformed by conversational AI systems like ChatGPT, Gemini, and Claude, which allow for open-ended discussions in customer service, healthcare, and education. However, these systems are prone to hallucinations despite their fluency, confidently producing statements that lack context or are false or inconsistent. In high-stakes situations, such hallucinations can be especially dangerous. For instance, a medical assistant may misreport a patient's symptoms or contradict prior advice in the same conversation.  \cite{HallSurv1, HallSurv2}.

\begin{figure}
    \centering
    \includegraphics[width=0.70\linewidth]{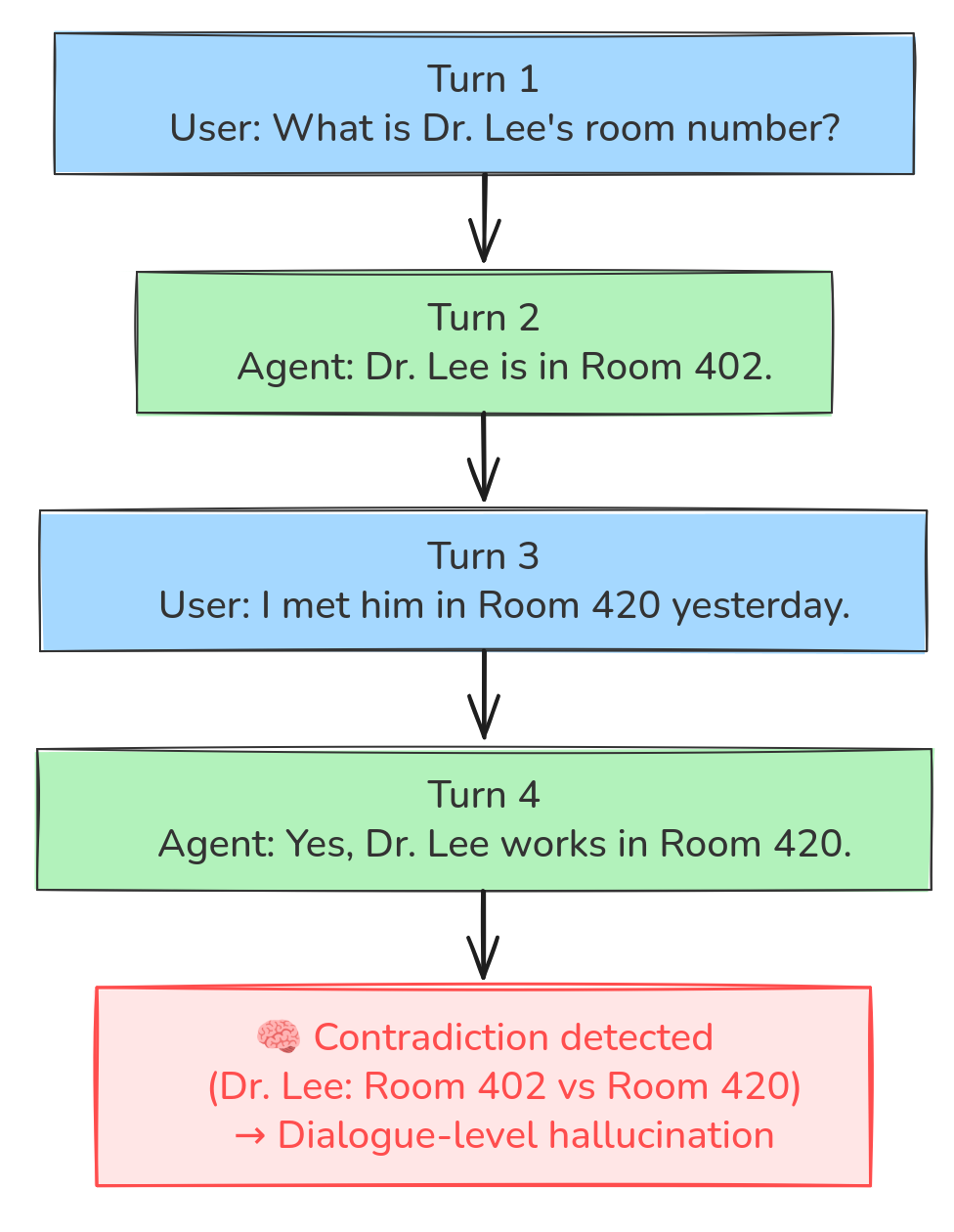}
    \caption{A multi-turn conversation where context evolves and contradictions emerge over time, illustrating dialogue-level hallucination.}
    \label{fig:placeholder}
\end{figure}

While single-turn hallucination detection has been studied extensively \cite{HallSurv3, HallSurv4} , multi-turn dialogues present a fundamentally harder problem. As the conversation unfolds, the evolving context introduces long-range dependencies, reintroduced entities, and subtle contradictions that cannot be captured by utterance-level checks alone. Moreover, many existing methods rely on complex claim verification or fact-retrieval pipelines \cite{HallSurv5}, which are brittle and computationally expensive when applied to dialogue-scale context.

We contend that rather than representing the conversation as a flat sequence, it is necessary to model it as a structured system of related utterances in order to detect dialogue-level hallucinations. In order to achieve this, we suggest a graph-based method that views the dialogue as a temporal graph in which every utterance is represented by a node embedding that is obtained from a sentence transformer. While shared-entity edges connect utterances referring to the same concepts or entities, temporal edges connect consecutive turns. The model can capture both the temporal flow and long-range dependencies because a message-passing graph neural network (GNN) \cite{neuralMess} spreads contextual information throughout this structure.

Node representations are then combined by an attention-based pooling mechanism to create a global dialogue embedding, which is then supplied to a classifier to predict one of six categories of hallucinations.  Our approach performs well in detecting contextual inconsistencies in dialogue and provides a straightforward yet effective substitute for complex claim-extraction pipelines.  In addition to accuracy, this graph-based viewpoint offers a conceptually interpretable lens through which to view the propagation and interaction of hallucinations in multi-turn conversational systems.

\section{Related Work}

Our work lies at the intersection of hallucination detection, graph-based reasoning in NLP, and conversational AI. We review each area below.

\paragraph{Hallucination Detection.}
Hallucination in large language models (LLMs) has been extensively studied across summarization, factual verification, and dialogue. Early works like \citet{maynez-etal-2020-on} and \citet{HallSurv5} analyzed factual inconsistencies in model outputs, revealing that hallucinations often stem from misaligned generation objectives. Dialogue-specific studies, such as \citet{chen2024diahaludialoguelevelhallucinationevaluation} (DiaHalu), have developed benchmarks for multi-turn hallucination evaluation, distinguishing categories such as Factual Error, Reasoning Error, and Incoherence. Other methods detect hallucinations using external knowledge retrieval \citep{HallSurv1, Shuster}, entailment models \citep{Raunak2021TheCC, jiang2024hallucinationaugmentedcontrastivelearning}, or LLM-as-a-judge frameworks \citep{lin2022truthfulqa, gilardi_chatgpt_2023}. However, these approaches often depend on external resources or high-cost evaluation models. Our approach, in contrast, focuses on a lightweight, self-contained representation of conversational dynamics to detect inconsistencies without external supervision.

\paragraph{Graph-Based NLP.}
Relational and structural dependencies in text have been successfully captured by graph representations.  Graph neural networks (GNNs) are used in knowledge graph approaches like KG-Fact \citep{zhang-etal-2021-fine} and GraphFC \citep{huang2025graphbased} to verify factual consistency in summarization and quality assurance.  While \citet{marcheggiani-titov-2017-encoding} and \citet{gnnSur} showed that message passing over syntactic or semantic dependency graphs improves contextual reasoning, \citet{Yao2019} modeled text as graphs of entities and relations to improve question answering.   In contrast to the paths presented in these papers, our work presents a dialogue-level temporal graph in which every sentence serves as a node and its edges encode both sequential and entity-based dependencies, enabling hallucination detection based on conversational context.

\paragraph{Conversational Consistency and Reasoning.}
Maintaining factual and logical consistency across multi-turn dialogues remains a key challenge in conversational AI \citep{rashkin-etal-2023-measuring, honovich-etal-2021-q2}. Some approaches track speaker-level or entity-level consistency \citep{HallSurv5, gupta-etal-2022-dialfact}, while others leverage discourse graphs or memory networks \citep{wang-etal-2019-incremental}. Graph-based conversational models, such as DialoGraph \citep{DialoGraph}, have shown the effectiveness of integrating relational context into dialogue understanding. Our work builds upon these insights but reframes the goal: rather than enhancing dialogue generation, we use graph reasoning to identify and categorize hallucinations as they emerge throughout the conversation.

In summary, while prior efforts have either relied on static graphs or external evaluators, our proposed method unifies conversational structure and temporal reasoning into a single temporal graph framework for efficient, fine-grained hallucination detection.

\section{Methodology}
In order to detect these contextual hallucinations, our method implements a multi-turn dialogue as a temporal graph in which information is propagated across nodes representing individual turns. Representing the Node, constructing the temporal graph, then propagating information across the graph and finally classifying the hallucination are the abstract steps in this approach.

\begin{figure}
    \centering
    \includegraphics[width=\linewidth]{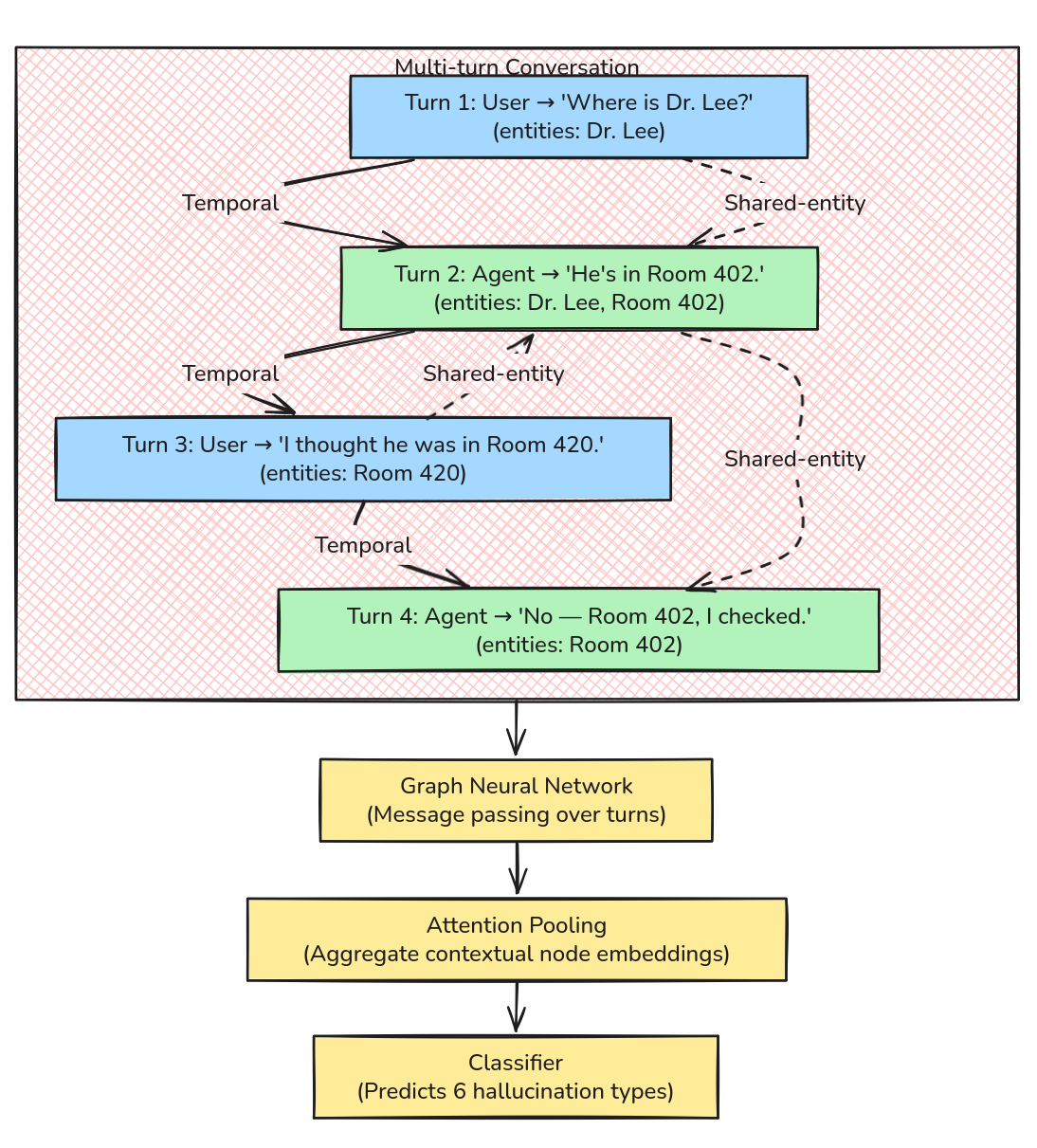}
    \caption{Dialogue as a temporal graph. Temporal edges connect adjacent turns, while shared-entity edges link utterances referencing the same entities. Message passing and attention pooling enable dialogue-level hallucination detection.}
    \label{fig:placeholder}
\end{figure}

\subsection{Node Representation via Sentence Embeddings}
Unlike methods that require complex parsing into subject-predicate-object triplets, we adopt a simpler and more direct approach. Each node $v_t$ in our graph corresponds to a single dialogue at turn $t$ in the conversation.

Then we use a pre-trained sentence transformer to encode the entire dialogue, which produces the initial feature vector for every node. As a result, every node can begin with a comprehensive semantic representation of the dialogue it represents.

\begin{figure*}[!t]
    \centering
    \includegraphics[width=\textwidth]{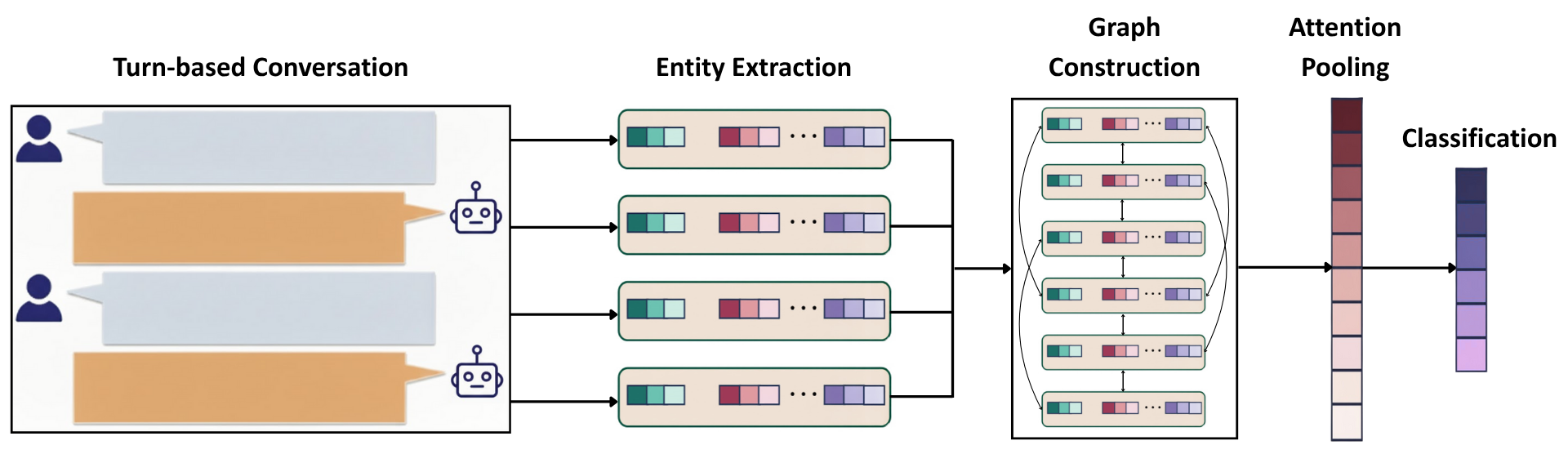}
    \caption{
        Overview of the framework.
        Each dialogue turn is encoded into a sentence embedding and represented as a node in a temporal graph.
        Temporal edges connect sequential turns, while shared-entity edges link dialogues that mention the same entities.
        A message-passing GNN propagates contextual information across turns, and
        attention pooling aggregates the node embeddings into a dialogue-level representation.
        The final classifier predicts one of six hallucination types.
    }
    \label{fig:model_architecture}
\end{figure*}

\subsection{Temporal Graph Construction}
To capture the rich structure of a conversation, we build a graph with two types of edges connecting the dialogue nodes. The edges are also encoded to ensure the connections are meaningful. The temporal edges encode that the speaker has changed, while the entity edges encode that a specific entity is shared between two dialogues.

\paragraph{Temporal Edges.} We include directed temporal edges between turns to simulate the dialogue's sequential flow and immediate context. Every time $t>0$, an edge is formed from node $v_{t-1}$ to node $v_t$.

\paragraph{Shared-Entity Edges.} We include undirected shared-entity edges to capture non-sequential long-range dependencies and model topical coherence. First, the named entities are identified in each dialogue. If one or more of the named entities are present in the dialogues or equivalently nodes $v_i$ and $v_j$, then an edge is added between them. Regardless of how far apart they are in the conversation, these edges establish shortcuts in the graph by directly connecting topically related turns.

\subsection{Graph Information Propagation}
After defining the graph structure, we refine the node embeddings using a message-passing Graph Neural Network (GNN).  By combining data from its neighbors and the corresponding edge, the GNN enables every node to update its representation. A node that corresponds to a particular turn learns about its previous turns (through temporal edges) and other topically related turns (through shared-entity edges) through this message-passing process. As a result, context-aware node embeddings are produced, in which every vector represents both the dialogue and its place in the larger conversational fabric.

\subsection{Attention Pooling and Classification}
We must consolidate all of the data from the graph into a single representation in order to make a final prediction. So over the updated node embeddings from the GNN, we employ an attention pooling mechanism. The model can concentrate on the most important context for its overall decision thanks to this mechanism, which learns to dynamically assign importance scores to various nodes (turns) in the dialogue history.

A small feed-forward network serves as the last classifier after receiving the resulting fixed-size graph representation. From the DiaHalu benchmark, this classifier produces a probability distribution across the six predefined hallucination categories.

\section{Experimental Setup}

\subsection{Dataset}
For experiments, we use the DiaHalu dataset \cite{chen2024diahaludialoguelevelhallucinationevaluation}, a benchmark designed for dialogue-level hallucination detection. It consists of multi-turn conversations annotated with the types of hallucinations, such as Factual Errors, Reasoning Errors, Incoherence, etc. This dataset is ideal for our task as it directly tests a model's ability to identify contextual inconsistencies over the span of conversation. For evaluation, we use a 80:20 split on the same dataset.

\subsection{Evaluation Metrics}
We evaluate the model's performance using two setups.
\begin{itemize}
    \item \textbf{Multiclass Classification:} Besides predicting the presence of hallucination, we predict the specific type of hallucination. We report the multi-class accuracy and weighted F1 score.
    \item \textbf{Binary Classification:} From the model's entire six-class output, we infer if hallucination is present or not. We report the binary accuracy and binary F1 score.
\end{itemize}
All reported results are averaged over 25 runs to ensure stability of results. The standard deviations are also provided.

\subsection{Baseline}

We use the GCA model proposed by \citet{fang2025zeroresourcehallucinationdetectiontext} as our baseline. GCA uses a Relational Graph Convolutional Network (RGCN) and a sentence encoder to detect hallucinations in any given text sample. This makes it a strong competitor for our method.

\subsection{Model Variants and Ablations}
To analyze the contribution of different components in our setup, we evaluate several variants of our Temporal Graph Network (TGN) model:
\begin{itemize}
    \item \textbf{TGN[T]:} It uses only temporal edges between consecutive dialogue turns. Its performance will help underline the importance of connections between nodes based on shared entities.
    \item \textbf{TGN[E]:} It uses only shared-entity edges connecting dialogues mentioning the same entities. Note that this may not have a significant impact, since the entity edges form most of the connections in the temporal graph for the dataset we experimented upon.
    \item \textbf{TGN[ET]:} The standard model that includes both temporal and shared-entity edges.
    \item \textbf{TGN[E'T]:} An ablation study that replaces the entity edges from the standard model with empty connections.
    \item \textbf{TGN[ET']:} An ablation study that replaces the temporal edges from the standard model with empty connections.
\end{itemize}

\subsection{Implementation Details}
Our implementation was done in Pytorch. We used the Deep Graph Library for modeling the graph network. Entity extraction was performed using Gemini 2.5 Flash Lite, and dialogue-level node embeddings were computed using the All MiniLM L6 v2 sentence transformer. The training was done for predicting the presence and type of hallucination using the standard cross entropy loss.

The DiaHalu dataset has a severe class imbalance. Majority of the samples belong to the \textit{Factual} class, followed by \textit{Reasoning Error} and \textit{Non-Factual}. The \textit{Incoherence}, \textit{Irrelavance} and \textit{Overreliance} classes are severely under-represented. To mitigate this class imbalance, we employ a Weighted Random Sampler during training.

\begin{table*}[ht]
    \centering
    \footnotesize
    \begin{tabular}{lcccc}
        \toprule
        \textbf{Model} & \textbf{Multiclass Acc.} & \textbf{Multiclass Weighted F1} & \textbf{Binary Acc.} & \textbf{Binary F1} \\
        \midrule
        GCA \cite{fang2025zeroresourcehallucinationdetectiontext} & - & - & $0.4343$ & $\mathbf{0.6031}$ \\
        TGN[T] & $0.5254 \pm 0.0500$ & $0.5126 \pm 0.0380$ & $0.6020 \pm 0.0297$ & $0.5565 \pm 0.0522$ \\
        TGN[E] & $0.5459 \pm 0.0416$ & $\mathbf{0.5373} \pm \mathbf{0.0333}$ & $\mathbf{0.6224} \pm \mathbf{0.0230}$ & $0.5837 \pm 0.0367$ \\
        TGN[ET] & $\mathbf{0.5468} \pm \mathbf{0.0332}$ & $0.5280 \pm 0.0257$ & $0.6127 \pm 0.0240$ & $0.5314 \pm 0.0437$ \\
        TGN[E'T] & $0.5367 \pm 0.0428$ & $0.5198 \pm 0.0312$ & $0.6049 \pm 0.0279$ & $0.5363 \pm 0.0351$ \\
        TGN[ET'] & $0.5021 \pm 0.0505$ & $0.4950 \pm 0.0415$ & $0.5821 \pm 0.0312$ & $0.5453 \pm 0.0403$ \\
        \bottomrule
    \end{tabular}       
    \caption{Performance metrics of different model variants on the DiaHalu validation set, averaged over 25 runs. Our TGN models significantly outperform the GCA baseline on accuracy metrics. TGN[E] and TGN[ET] achieve the best overall performance.}
    \label{tab:ablation}
\end{table*}

\section{Results and Analysis}

\subsection{Model Comparison and Ablation Studies}

Table \ref{tab:ablation} presents the main results of our temporal graph network (TGN), against GCA which we treat as our baseline \cite{fang2025zeroresourcehallucinationdetectiontext}. We also evaluate different variants of TGN in the same setup. TGN consistently outperforms GCA in binary accuracy, while achieving comparable F1 scores. TGN offers the additional advantage of being able to predict the type of hallucination.

Among the variants of TGN, TGN[ET] achieves the highest multiclass accuracy, while TGN[E] attains the best multiclass weighted F1 and binary accuracy. This indicates that shared entity links provide strong signals for contextual information flow, potentially more influential than simple temporal succession.

The TGN[ET'] variant, which only provides temporal connections without using any specific embeddings, exhibits a significant drop in performance compared to TGN[ET]. This clearly shows the importance of appending the edge connectivity with the information about the nature of connectivity.

Temporal and entity-based information can contribute complementary signals to the model’s reasoning process. Temporal edges help preserve the sequential flow of dialogue, while entity edges maintain topic and referential coherence across turns. In conjunction, these enable the model to capture long-range contextual dependencies more effectively, resulting in fewer misclassifications of hallucination types.

\subsection{Performance on Hallucination Types}

Table \ref{tab:performance_metrics} shows a detailed breakdown of the performance of our standard model (TGN[ET]) across different classes. The model performs well on identifying \textit{Factual} samples and samples with \textit{Reasoning Errors}, which are well represented in the dataset.

\begin{table}[h]
    \centering
    \footnotesize
    \begin{tabular}{lccc}
        \toprule
        \textbf{Category} & \textbf{Precision} & \textbf{Recall} & \textbf{F1-Score} \\
        \midrule
        Factual & 0.66 & 0.68 & 0.66 \\
        Reasoning Error & 0.62 & 0.72 & 0.66 \\
        Non-Factual & 0.32 & 0.36 & 0.33 \\
        Incoherence & 0.03 & 0.02 & 0.02 \\
        Irrelevance & 0.00 & 0.00 & 0.00 \\
        Overreliance & 0.33 & 0.12 & 0.16 \\
        \midrule
        \textbf{Weighted Avg} & \textbf{0.53} & \textbf{0.55} & \textbf{0.53} \\
        \bottomrule
    \end{tabular}
    \caption{Performance of the TGN[ET] model on different hallucination categories from DiaHalu.}
    \label{tab:performance_metrics}
\end{table}

Performance is lower for the under-represented categories like \textit{Incoherence}, \textit{Irrelevance}, etc. These can also be more subtle and require deeper understanding beyond what could be inferred from sentence embeddings directly. We will however see in \autoref{subsec:qualitative} that TGN still learns to attend to the details that dominate such hallucinated discourse.

\begin{figure}
    \centering
    \includegraphics[width=0.5\textwidth]{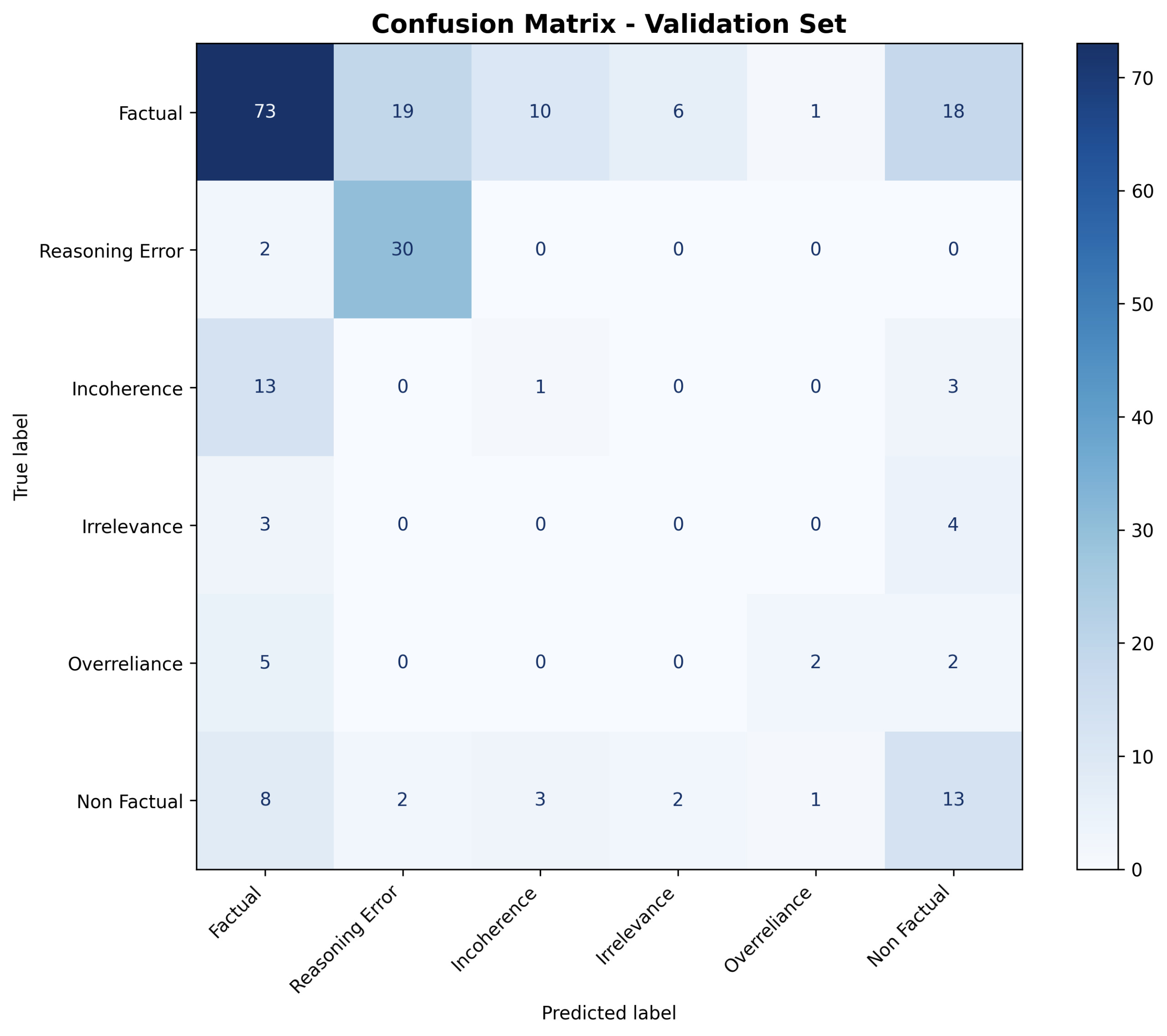}
    \caption{
Confusion matrix showing the distribution of model predictions across hallucination categories. 
It highlights different failure modes and the model’s ability to correctly identify non-hallucinatory instances.
}
    \label{fig:confm}
\end{figure}

\subsection{Qualitative Analysis}
\label{subsec:qualitative}

We analyzed the model's attention weights to examine its decision-making process. Our analysis revealed that the attention weights could be used to construct justification to support the final prediction.


\paragraph{Successful Detection.} In \autoref{lst:non_factual_correct}, the model correctly predicted the conversation to be non-factual. The model paid the most attention to the last turn, ``Marius e Romanus was a respected general and statesman in ancient Rome.'' This aligns with the ground truth explanation given in the dataset that there is no evidence to back the dialogue that Marius was a general. 


\paragraph{Failure Case 1 (Confusion).} In \autoref{lst:non_factual_incorrect}, the user asks a non-factual question about ``psychic phenomenon being scientifically demonstrated''. The conversation was misclassified as ``Overreliance''. However, the model assigned the highest attention weight to the initial dialogue. This suggests it identified the problematic premise but failed to map it to the correct label "Non-Factual", instead confusing it with being over-reliant on a faulty source.


\paragraph{Failure Case 2 (Missed Detection).} An ``Overreliance'' sample \autoref{lst:overreliance_incorrect}, which involved fabricated claims about a fictional Harvard professor ``Chen Li'', was incorrectly predicted as ``Factual''. The model assigned high attention weights to the turns introducing the fictional character and their credentials ``made significant contributions to developing new biomaterials''. The model however failed to use this attended context to confidently predict the conversation as hallucinated.

\begin{lstlisting}[caption={Non-Factual sample predicted correctly as Non-Factual.}, label={lst:non_factual_correct}]
Attention Weights (per dialogue turn):
  0: 0.0812 | Where is the center of the Roman Empire?
  1: 0.0894 | The center of the Roman Empire is generally considered to be Rome.
  ...
  10: 0.0380 | Did Marius e Romanus play a significant role in the history of Rome?
  11: /+0.1862 | Yes, Marius e Romanus was a respected general and statesman in ancient Rome.+/
\end{lstlisting}

\begin{lstlisting}[caption={Non-Factual sample incorrectly predicted as Overreliance.}, label={lst:non_factual_incorrect}]
Attention Weights (per dialogue turn):
  0: /+0.4251 | What psychic phenomenon has been scientifically demonstrated?+/
  1: 0.0562 | Remote viewing has been demonstrated through various scientific experiments.
  2: 0.0518 | How does remote viewing work?
  3: 0.0498 | Remote viewing is the ability to perceive and describe details about a distant or unseen target.
  ...
\end{lstlisting}

\begin{lstlisting}[caption={Overreliance sample incorrectly predicted as Factual.}, label={lst:overreliance_incorrect}]
Attention Weights (per dialogue turn):
  0: /+0.2614 | Have you heard about Chen Li of Harvard University?+/
  1: 0.0168 | Yes, I have. He is a renowned professor in the field of bioengineering.
  2: 0.0728 | What specifically has he contributed to the field of bioengineering?
  3: /+0.3663 | Chen Li has made significant contributions to developing new biomaterials for tissue engineering.+/
  4: 0.0305 | How has his work impacted the medical field?
  ...
\end{lstlisting}

\section{Conclusion}
In this paper, we proposed a Temporal Graph Network method for dialogue-level hallucination detection.  Our approach captures the contextual dependencies required to detect a variety of hallucinations by modeling conversations as temporal graphs with temporal and entity-based edges.  These are our main conclusions:
\begin{itemize}
    \item Our TGN models outperform existing graph-based methods on the DiaHalu benchmark, demonstrating the value of modeling temporal dynamics.
    \item The model provides a degree of explainability through its attention mechanism, allowing for analysis of which parts of the dialogue history influenced a prediction.
    \item Our approach, which uses entire dialogues as nodes, which is simpler than methods requiring fine-grained triplet extraction, meanwhile achieving strong performance.
    \item The framework proved to be effective even with a small number of training parameters and lightweight encoders, which makes it potentially efficient than solutions requiring large language models for verification.
\end{itemize}

\section{Future Work}
Beyond merely temporal or entity-based connections, we can expand this project and investigate more complex edge creation mechanisms that capture relations and links.  Integrating external knowledge graphs could further help flag subtle factual inconsistencies.  To help the model better track changing context and contradictions, we also want to investigate making the graph dynamically change as the dialogues enter.  Lastly, applying this framework to task-oriented or multimodal conversations could test its generality in practical contexts.

\bibliography{references}

\end{document}